\documentclass{article}
\usepackage{ijcai21}

\usepackage{times}
\usepackage{soul}
\usepackage{url}
\usepackage[hidelinks]{hyperref}
\usepackage[utf8]{inputenc}
\usepackage[small]{caption}
\usepackage{graphicx}
\usepackage{amsmath}
\usepackage{amsthm}
\usepackage{booktabs}
\usepackage{algorithm}
\usepackage{algorithmic}
\newcommand\citeay[1]{\citeauthor{#1}~[\citeyear{#1}]}

\title{A Survey on Low-Resource Neural Machine Translation}

\author{
Rui Wang\and
Xu Tan\and
Renqian Luo\and
Tao Qin\And
Tie-Yan Liu\\
\affiliations
Microsoft Research Asia\\
\emails
\{ruiwa, xuta, t-reluo, taoqin, tyliu\}@microsoft.com
}

\begin{document}
\maketitle






\begin{abstract}
Neural approaches have achieved state-of-the-art accuracy on machine translation but suffer from the high cost of collecting large scale parallel data. Thus, a lot of research has been conducted for neural machine translation (NMT) with very limited parallel data, i.e., the low-resource setting. In this paper, we provide a survey for low-resource NMT and classify related works into three categories according to the auxiliary data they used: (1) exploiting monolingual data of source and/or target languages, (2) exploiting data from auxiliary languages, and (3) exploiting multi-modal data. We hope that our survey can help researchers to better understand this field and inspire them to design better algorithms, and help industry practitioners to choose appropriate algorithms for their applications. 
\end{abstract}

\section{Introduction}
Machine translation (MT) automatically translates from one language to another without human labor, which brings convenience and significantly reduces the labor cost in international exchange and cooperation. Powered by deep learning, neural machine translation (NMT)~\cite{bahdanau2015neural,vaswani2017attention} has become the dominant approach for machine translation. Compared to conventional rule-based approaches and statistical machine translation (SMT), NMT enjoys two main advantages. First, it does not require professional human knowledge and design on translation perspective (e.g., grammatical rules). Second, neural network can better capture the contextual information in the entire sentence, and thus conduct high quality and fluent translations.

One limitation of NMT is that it needs large scale of parallel data for model training. While there are thousands of languages in the world\footnote{https://en.wikipedia.org/wiki/Language}, major popular commercial translators (e.g., Google translator, Microsoft translator, Amazon translator) only support tens or a hundred languages because of the lack of large-scale parallel training data for most languages. To handle those languages with limited parallel data, many algorithms have been designed for low-resource NMT in recent years. Therefore, a review on low-resource NMT is very helpful for fresh researchers entering this area and industry practitioners. Although there already exists surveys on many aspects of NMT (e.g., domain adaptation~\cite{chu2018survey}, multilingual translation~\cite{dabre2020survey}), a comprehensive survey for low-resource NMT is still missing. Therefore, in this paper, we conduct a comprehensive and well-structured survey on low-resource NMT to fill in this blank. 

\begin{figure}[t] 
\centering
\includegraphics[width=2.5in]{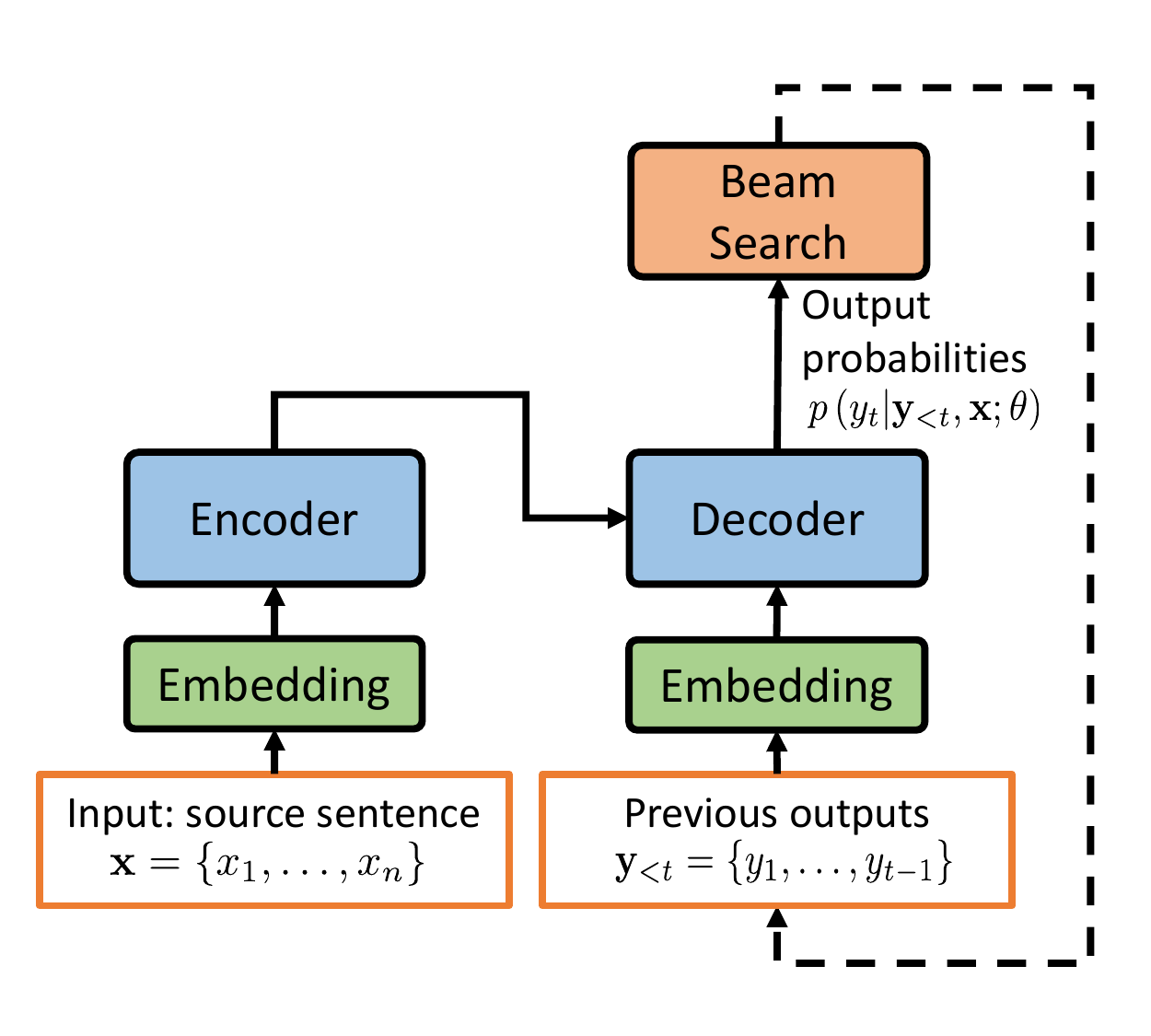} 
\caption{NMT model}
\label{fig:NMT} 
\end{figure} 

\paragraph{NMT basics.} An NMT model $\theta$ translates a sentence $\mathbf{x}$ in the source language to a sentence $\mathbf{y}$ in the target language. With a parallel training corpus $\mathbf{C}$, the model $\theta$ is trained by minimizing the negative log-likelihood loss:
\begin{equation}
    L_{\theta} = \sum\nolimits_{(\mathbf{x},\mathbf{y}) \in \mathbf{C}} -\mbox{log} p \left( \mathbf{y} | \mathbf{x};\theta\right).
\end{equation}
As shown in Fig. \ref{fig:NMT}, NMT models are commonly auto-regressive and generate the target sentence from left to right. Considering that $\mathbf{y}$ contains $m$ words, the conditional probability $p \left( \mathbf{y} | \mathbf{x};\theta\right)$ can be written as:
\begin{equation}
    p \left( \mathbf{y} | \mathbf{x};\theta\right) = \Pi_{t=1}^{m} p\left(y_t|\mathbf{y}_{<t},\mathbf{x};\theta\right).
\end{equation}
The encoder-decoder framework is widely used in NMT, where the encoder converts the source sentence into a sequence of hidden representations and the decoder generates target words conditioned on the source hidden representations and previously generated target words. The encoder and decoder can be recurrent neural networks ~\cite{dong2015multi}, convolutional neural networks~\cite{gehring2017convolutional}, and Transformer~\cite{vaswani2017attention}. In the inference stage, beam search is usually used to generate the target sentence based on the decoder output.

\begin{figure*}[t] 
\centering
\includegraphics[width=6in]{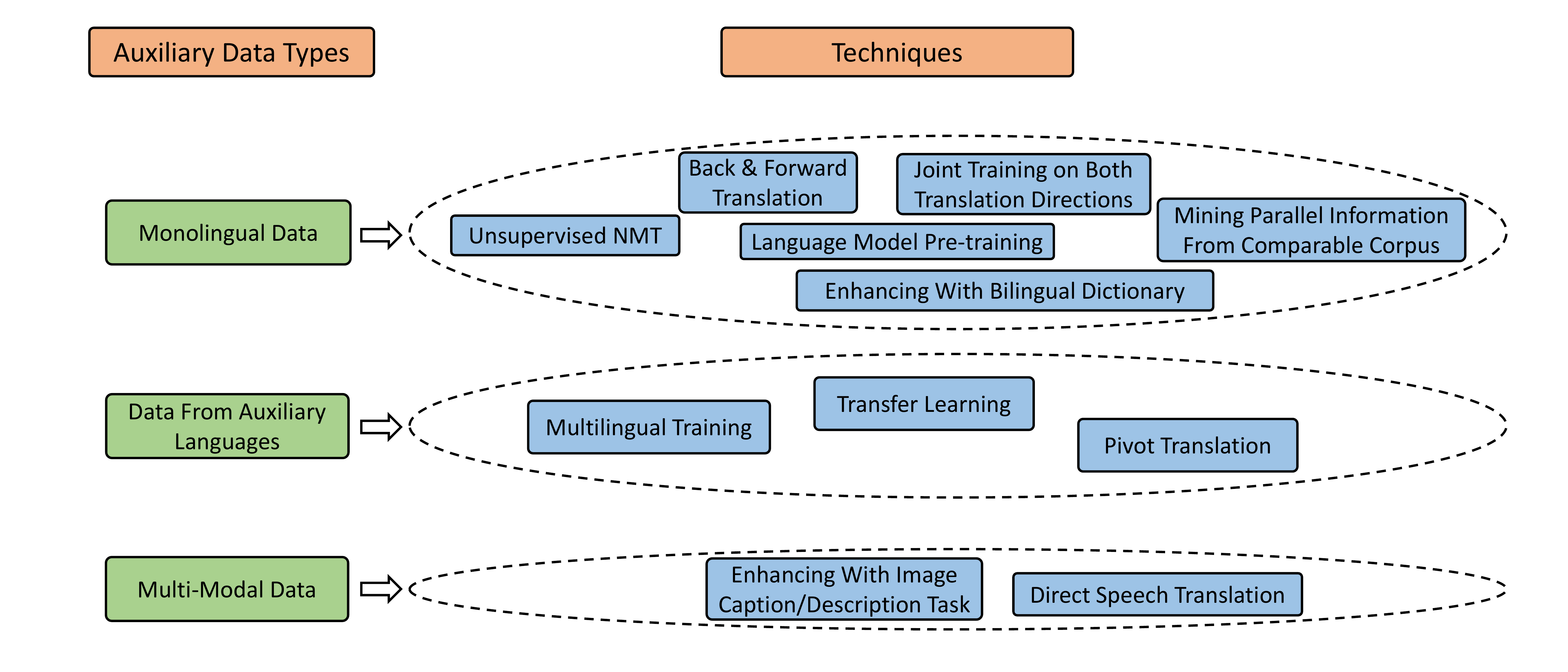} 
\caption{Overview}
\label{fig:Overview} 
\end{figure*} 

\paragraph{Organization of this survey.} Due to the lack of parallel sentence pairs, leveraging data other than parallel sentences is essential in low-resource NMT. In this paper, as shown in Fig. \ref{fig:Overview}, we categorize existing algorithms on low-resource NMT into three categories according to the data they use to help a low-resource language pair: 
\begin{itemize}
    \item \emph{Monolingual data.} Leveraging unlabeled data to boost machine learning models is a popular and effective approach in various areas. Similarly, in NMT, leveraging unlabeled monolingual data attracts lots of attentions (see Section~\ref{sec:mono}) since collecting monolingual data is much easier and of lower cost than parallel data.
    \item \emph{Data from auxiliary languages.} Languages with similar syntax and/or semantics are helpful to each other when training NMT models. Leveraging data from related and/or rich-resource languages has shown great success in low-resource NMT (see Section~\ref{sec:multi}). 
    \item \emph{Multi-modal data.} Multi-modal data (e.g., parallel data between text and image) has also been used in low-resource NMT, as reviewed in Section~\ref{sec:multimodal}.
\end{itemize}
In addition to reviewing algorithms, we also summarize widely used data corpora for low-resource NMT in Section~\ref{sec:data}. We conclude this survey and discuss future research directions in the last section.

\section{Exploiting Monolingual Data}
\label{sec:mono}

\begin{figure*}[t] 
\centering
\includegraphics[width=7in]{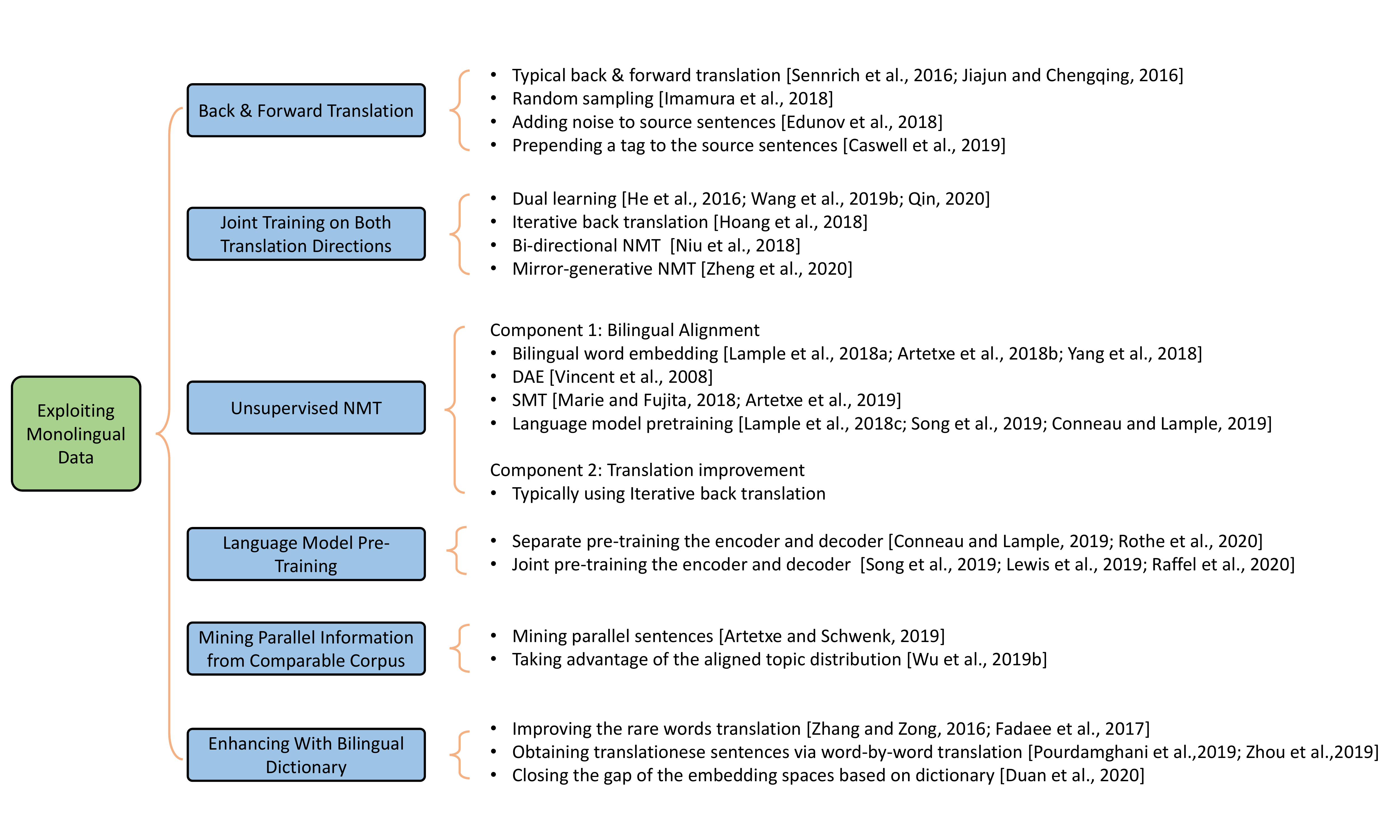} 
\caption{Overview of works exploiting monolingual data}
\label{fig:Monolingual} 
\end{figure*} 

Monolingual data contains a wealth of linguistic information (e.g., grammar and contextual information) and is more abundant and easier to obtain than bilingual parallel data, which is useful to improve the translation quality especially in low-resource scenario. Plenty of works have exploited monolingual data in NMT systems, which we categorize into several aspects: (1) back translation, which is a simple and promising approach to take advantage of the target-side monolingual data~\cite{BT}, (2) forward translation also called knowledge distillation, which utilizes the source-side monolingual data~\cite{FT}, (3) joint training on both translation directions, which can take advantage of the monolingual data on both the source and target sides~\cite{he2016dual,hoang2018iterative,niu2018bi,zheng2020mirror}, (4) unsupervised NMT, which builds NMT models with only monolingual data, and can be applied to the language pairs without any parallel data~\cite{lample2018unsupervised,artetxe2018unsupervised}, (5) pre-training, which leverages monolingual data with self-supervised training for language understanding and generation, and thus improves the quality of NMT models~\cite{conneau2019cross,song2019mass,lewis2019bart}, (6) comparable monolingual corpus, which contains implicit parallel information and can improve the translation quality~\cite{wu2019extract}, and (7) enhancing with bilingual dictionary, where the bilingual dictionary is used together with monolingual data to enhance the translation on low-resource languages. In this section, we provide an overview of these methods on exploiting monolingual data in NMT.

\subsection{Back \& Forward Translation}

In back translation, pseudo parallel sentence pairs are generated by translating the target-side monolingual sentences to the source language via a translation system in the reverse direction~\cite{BT}, while in forward translation, pseudo parallel sentence pairs are generated by translating the source-side monolingual sentences to the target language via a translation system in the same direction~\cite{FT}. Then, the pseudo parallel data is mixed with the original parallel data to train an NMT model. It has been shown that back and forward translation provides promising performance gain on NMT systems~\cite{BT,FT,poncelas1804investigating}.

Besides the typically used beam search~\cite{BT,FT}, there are also some other methods to generate the pseudo parallel data: (1) random sampling according to the output probability distribution~\cite{imamura2018enhancement}, (2) adding noise to source sentences generated by beam search~\cite{edunov2018understanding}, and (3) prepending a tag to the source sentences generated by beam search~\cite{caswell2019tagged}. It is observed that random sampling and adding noise only works well on high resource setting compared to standard beam search~\cite{edunov2018understanding}, while prepending a tag performs the best on both high and low resource settings~\cite{caswell2019tagged}. In addition, a mixed  pseudo parallel data generated by BT and copying the target monolingual sentences as the source sentences can further improve the translation quality on low-resource languages~\cite{currey2017copied}.

\subsection{Joint Training on Both Translation Directions}
\label{sec:both-side}

Considering that both the source and target sides monolingual data has valuable information, some works leverage both of them via joint training on the two translation directions. Dual learning~\cite{he2016dual,qin2020dual} simultaneously improves the two models on both translation directions by aligning the original monolingual sentences $x$ and the sentences $x’$ translated forward and then backward ($x\to y’\to x’$) by the two models. \citeay{wang2019multi} further improve dual learning by introducing multi-agent for both translation directions. Intuitively, a better reverse translation model leads to better back-translation sentences, and thus leading to a better NMT system. Iterative back translation~\cite{hoang2018iterative} simultaneously trains the NMT models on both translation directions and iteratively updates the back-translated corpus via the updated better NMT models. Bi-directional NMT~\cite{niu2018bi} trains both the translation directions in the same model with a tag indicating the direction at the beginning of source sentences, and then leverages both source-side and target-side monolingual data by back and forward translation. Mirror-generative NMT~\cite{zheng2020mirror} jointly trains the translation models on both directions and the language models for both source and target languages with a shared latent variable.


\subsection{Unsupervised NMT}
To deal with the zero-resource translation scenario without any parallel sentences, a common approach is unsupervised learning for NMT~\cite{lample2018unsupervised,artetxe2018unsupervised}, which typically relies on two components to ensure the learning efficiency and quality: (1) bilingual alignment, which enables the model with good alignments between the two languages, and (2) translation improvement, which gradually improves the translation quality by iterative learning, typically through back translation~\cite{BT,hoang2018iterative,zhang2018joint}.

\paragraph{Bilingual alignment.} How to initially align between the two languages is an open problem. There are mainly four kinds of approaches: (1) bilingual word embedding~\cite{mikolov2013distributed,artetxe2017learning,zhang2017adversarial,bojanowski2017enriching,lample2018word}, where the NMT system can either start from a word-by-word translation derived from the bilingual word embedding~\cite{lample2018unsupervised} or initialize the embedding parameters according to bilingual word embedding~\cite{artetxe2018unsupervised,yang2018unsupervised}, (2) denoising auto-encoder (DAE)~\cite{vincent2008extracting}, which can build a shared latent space of two languages by learning to reconstruct sentences in the two languages from a noised version~\cite{lample2018unsupervised,artetxe2018unsupervised,yang2018unsupervised}, (3) unsupervised statistical machine translation (SMT), where an initial alignment can be obtained through the back-translated corpora generated by an unsupervised SMT system~\cite{marie2018unsupervised,artetxe2019effective}, and (4) language model pre-training~\cite{lample2018phrase,song2019mass,conneau2019cross,ren2019explicit}, which is discussed in detail in Section \ref{sec:Pre-training}.




\paragraph{Translation improvement.} The translation quality need to be further improved based on the initial alignment, where iterative back translation is commonly used~\cite{lample2018unsupervised,lample2018phrase,song2019mass}. Some works study on improving the iterative back translation process in unsupervised NMT. Filtering out low-quality pseudo parallel sentence pairs is one strait-forward and effective method \cite{khatri2020filtering}. \citeay{sun2019unsupervised,sun2020unsupervised} propose to add a term in the training objective to avoid forgetting the alignment from bilingual word embedding during the interactively training. Moreover, unsupervised SMT can also be utilized to boost the iterative back translation. One approach is to first construct pseudo parallel data by leveraging both the unsupervised SMT and NMT systems for back translation and then train the NMT models with the pseudo parallel data~\cite{lample2018phrase,marie2019nict,ren2020retrieve}. In addition, SMT can also act as a posterior regularization to denoise the pseudo parallel data generated by NMT systems~\cite{ren2019unsupervised}.


\subsection{Language model pre-training}
\label{sec:Pre-training}

Leveraging monolingual data to pre-train language models is effective for many language understanding and generation tasks~\cite{devlin2018bert}. Since NMT requires the capability of both language understanding (e.g., NMT encoder) and generation (e.g., NMT decoder), pre-training language model can be very helpful for NMT, especially low-resource NMT. Previous works on language model pre-training for NMT can be divided into two categories depending on the encoder and decoder in NMT are pre-trained separately or jointly. We then review the works according to the two categories. 

\paragraph{Separate pre-training.} Some works pre-train the encoder or/and the decoder separately. \citeay{ramachandran2016unsupervised} first separately initialize the encoder and decoder with language models proposed by \citeay{jozefowicz2016exploring}, and then fine-tune with supervised parallel data. XLM~\cite{conneau2019cross} initialize the encoder and decoder with separate language models training by a combination of masked language modeling (MLM)~\cite{devlin2018bert}, where some tokens in the text are masked and the model learns to predict the masked tokens, and translation language modeling (TLM), which extends MLM by concatenating parallel sentence pairs as the input sentences. \citeay{rothe2020leveraging} investigate to initialize the encoder and decoder with variant models, including BERT~\cite{devlin2018bert}, GPT-2~\cite{radford2018improving,radford2019language}, RoBERTa~\cite{liu2019roberta} and random initialization. It is observed the best performance on English-Germany by a model with BERT-initialized encoder and randomly initialized decoder, or a model with shared encoder and decoder initialized with BERT. After initialing with separately pre-trained encoder and decoder, \citeay{varis2019unsupervised} introduce Elastic Weight Consolidation~\cite{kirkpatrick2017overcoming} into fine-tuning in order to avoid forgetting the language models. \citeay{zhu2019incorporating} fuse the representations extracted by BERT to the encoder and decoder via attention mechanisms. A drawback of separately pre-training encoder and decoder is that it cannot well train the encoder-decoder-attention, which is very important in NMT to connect the source and target representations for translation. Therefore, some works propose to jointly pre-train the encoder, decoder and attention for better translation accuracy.

\paragraph{Joint pre-training.} In order to simultaneously learn to understand the input sentences and improve the language generation capability, as well as jointly pre-train each component in NMT models (encoder, decoder and encoder-decoder-attention), MASS~\cite{song2019mass} proposes masked sequence to sequence learning that randomly masks a fragment (several consecutive tokens) in the input sentence of the encoder, and predicts the masked fragment in the decoder. Later, BART~\cite{lewis2019bart} proposes to add noises and randomly mask some tokens in the input sentences in the encoder, and learn to reconstruct the original text in the decoder. T5~\cite{raffel2020exploring} randomly masks some tokens and replace the consecutive tokens with a single sentinel token.

\subsection{Exploiting Comparable Corpus}
Monolingual data of different languages that refer to the same entity (e.g., English and Chinese Wikipedia pages that describe the same object) can be regarded as comparable corpus, which is easier to be obtained compared to parallel data and contains implicit parallel information for NMT systems. The challenge is how to mine the parallel sentences from the comparable corpus and some approaches are proposed to solve this problem. LASER~\cite{artetxe2019massively} is a toolkit based on cross-lingual sentence embeddings, which is a good choice to mine parallel data~\cite{schwenk2020wikimatrix,schwenk2019ccmatrix}. \citeay{wu2019extract} propose to first extract potential aligned target sentences given a source sentence, and then make the target sentences better aligned with the source sentence by revising them via an editing mechanism. A self-supervised learning method is proposed in~\cite{ruiter2019self}, where finding semantically aligned sentences is considered as an auxiliary task for translation. Besides mining parallel sentence pairs, \citeay{wu2019machine} take advantage of the aligned topic distribution for weakly paired documents, which is suitable for documents related to the same event or entity but not implicitly aligned in sentences.

\subsection{Enhancing With Bilingual Dictionary}

The bilingual dictionary of a low-resource language pair can be collected either by human annotation or word embedding based alignment~\cite{sennrich2015neural,conneau2017word,zhang2017earth,artetxe2018robust,zhang2017adversarial}, which is much easier to obtain than the bilingual parallel sentences. Since the bilingual dictionary contains only word-level information, it is usually used with monolingual data to improve the translation. Existing works utilizing the bilingual dictionary can be categorized into three ways. First, bilingual dictionary is used to improve the rare words translation. \citeay{zhang2016bridging} build pseudo parallel sentences by translating source-side monolingual sentences (that contain rare words) to target language via SMT (that is built based on the bilingual dictionary). \citeay{fadaee2017data} augment the parallel data by replacing some words in parallel sentences with rare words. Second, bilingual dictionary can also be used to perform word-by-word translation on monolingual data, and accordingly help to improve the low-resource NMT. \citeay{pourdamghani2019translating} propose a two-step approach, which first translates the source monolingual sentence to translationese sentence word-by-word using bilingual dicrionary, and then trains a translation model on translationese-to-target. \citeay{zhou2019handling} augment the parallel training data by first re-ordering monolingual sentences in the target language to match the source language and then obtaining pseudo source sentences via word-by-word translation. Third, a recent study~\cite{duan2020bilingual} propose to close the gap of the embedding spaces between the source and target languages by establishing anchoring points based on dictionary, which can help to build NMT models based on only dictionary and monolingual data.




\subsection{Summary and Discussions}

Back/forward translation and the joint training on both translation directions utilize monolingual data to improve translation models. Unsupervised NMT uses only monolingual data to get an initial alignment and improve the translation via iterative back translation. Language model pre-training initializes the NMT models with language understanding and generation capability using only monolingual data. Comparable corpora are strong supplements to parallel corpus, from which parallel sentences can be extracted based on language models or translation models. Bilingual dictionary contains word-level parallel information, which is helpful on the alignment between two languages. The above techniques can be combined with each other to gain more in low-resource NMT. For example, back/forward translation and the joint training methods on both translation directions can be applied to any existing translation models, and thus can be easily combined with other techniques in low-resource NMT. Moreover, the pre-trained model can either be fine-tuned to translation task via parallel data that may be extracted from comparable corpora, or used as an initial model for unsupervised NMT.

\section{Exploiting Data From Auxiliary Languages}
\label{sec:multi}

\begin{figure*}[t] 
\centering
\includegraphics[width=7in]{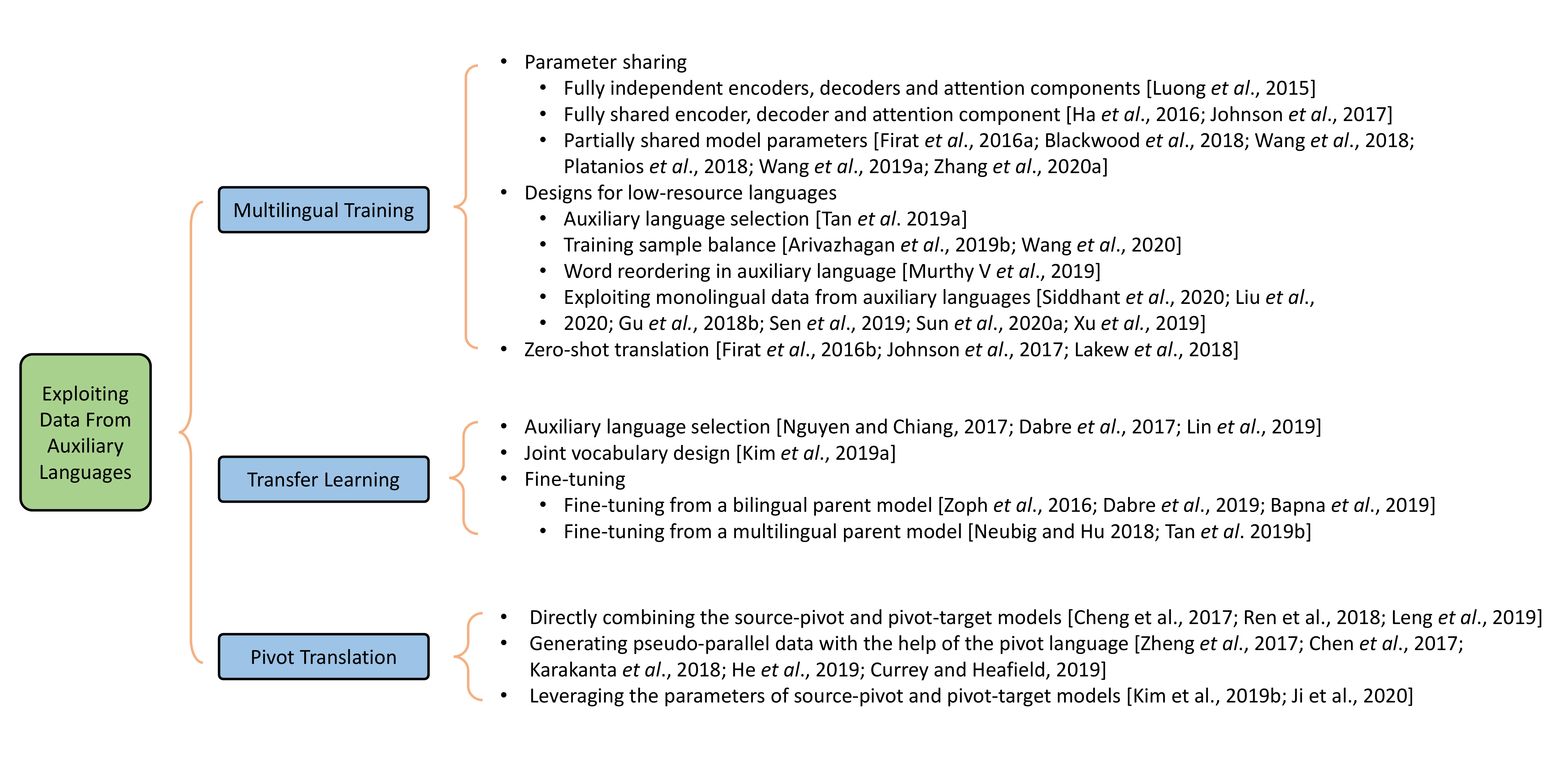} 
\caption{Overview of works exploiting data from auxiliary languages}
\label{fig:Multilingual} 
\end{figure*} 

Human languages share similarities with each other in several aspects: (1) languages in the same/similar language family or typology may share similar writing script, word vocabulary, word order and grammar, (2) languages can influence each other, and a foreign word from another language can be incorporated into a language as it is (referred as loanword). Accordingly, corpora of related languages can be exploited to assist the translation between a low-resource language pair~\cite{dabre2020survey}. The methods to leverage multilingual data into low-resource NMT can be categorized into several types: (1) multilingual training, where the low-resource language pair is jointly trained with other language pairs in one model~\cite{johnson2017google}, (2) transfer learning~\cite{zoph2016transfer}, where a parent NMT model usually containing rich-resource language pairs is first trained and then fine-tuned on low-resource language pair, and (3) pivot translation, where one or more pivot languages are selected as a bridge between the source and target languages and in this way the source-pivot and pivot-target data can be exploited to help the source-target translation. In the following subsections, we introduce the works in each category, respectively.

\subsection{Multilingual training}

Multilingual training enjoys three main advantages. First, training multiple language pairs in a single model through parameter sharing can significantly reduce the cost of model training and maintenance compared with training multiple separate models, and can collectively learn the knowledge from multiple languages to help low-resource languages. Second, low-resource language pairs benefit from related rich-resource languages pairs through joint training. Moreover, multilingual NMT offers the possibility to translate on language pairs that are unseen during training, which is called zero-shot translation. In the following paragraphs, we summarize the works on multilingual training from three perspectives (i.e., parameter sharing, designs for low-resource languages and zero-shot translation).

\paragraph{Parameter sharing.} There are different ways to share model parameters in multilingual training. First, all the encoder, decoder and attention components are independent among different languages~\cite{luong2015multi,dong2015multi,zoph2016multi}. Second, fully shared encoder, decoder and attention components are considered across languages, where a language-specific token is added in the source sentence to specify the target language~\cite{ha2016toward,artetxe2019massively,johnson2017google,tan2019multilingual2}. Third, in order to simultaneously exploit the characteristic and commonality of different languages, as well as keeping the model compact, some works consider to partially share the model parameters. \citeay{firat2016multi} adopt a shared attention mechanism and language-specific encoders and decoders. \citeay{blackwood2018multilingual} propose to use a specific attention mechanism in the decoder for each target language and share all the remaining model parameters, which is shown to improve the word alignments. \citeay{sachan2018parameter} propose to partially share the attention mechanism. \citeay{wang2018three} focus on one-to-many scenario and consider to use language-dependent positional embeddings and partially share the hidden layers in the decoder. \citeay{platanios2018contextual} introduces a contextual parameter generator for each language pair, which generates the parameters of the encoder and decoder based on the source and target language embeddings. \citeay{wang2019compact} improves the translation quality by using language-sensitive embeddings and attentions, as well as incorporating language-sensitive discriminators in the decoder. \citeay{zhang2020improving} introduce a linear transformation between the shared encoder and decoder for each target language, which requires only one more weight matrix for an additional target language.


\paragraph{Designs for low-resource languages.} To better exploit the knowledge from multiple languages to help low-resource languages, a lot of works design to improve the multilingual training from different aspects:   
\begin{itemize}
    \item \emph{Auxiliary language selection.} How to effectively select and utilize auxiliary languages is critical to improve the performance of low-resource language pairs in multilingual NMT. Most works consider to select rich-resource languages in the same language family as auxiliary languages, and achieve significant improvement~\cite{gu2018universal,guzman2019flores,neubig2018rapid}. \citeay{tan2019multilingual}  propose to cluster the languages based on language embedding, which shows better performance than clustering by language family. \citeay{wang2019target} focus on translating low-resource languages to English with the help of a target conditioned sampling algorithm, where a target sentence is sampled and the source sentences from all the corresponding parallel sentences in multiple languages are chosen based on language-level and sentence-level similarity. 
    \item \emph{Training sample balance.} Considering the limited model capacity and the various training data sizes among different languages, the model may have a bias to rich-resource languages. Accordingly, balancing the data sizes is important for low-resource languages in multilingual NMT. Temperature based sampling is one promising approach, where the temperature term needs to be manually chosen~\cite{arivazhagan2019massively}. \citeay{wang2020balancing} propose a method to automatically weight the training data sizes.
    \item \emph{Word reordering in auxiliary language.} Pre-ordering the words in auxiliary language sentences to align with the desired low-resource language also brings benefits to low-resource NMT~\cite{murthy2019addressing}.
    \item \emph{Monolingual data from auxiliary languages} can also be utilized to improve the low-resource languages by introducing back translation~\cite{BT}, cross-lingual pre-training~\cite{siddhant2020leveraging,liu2020multilingual} and meta-learning~\cite{gu2018meta} in multilingual model. Furthermore, multilingual NMT can also be trained with monolingual data only by extending the unsupervised NMT to multilingual model~\cite{sen2019multilingual,sun2020knowledge}, or aligning the translations to the same language via different paths in a multilingual model~\cite{xu2019polygon}.
\end{itemize}

\paragraph{Zero-shot translation.} Multilingual training brings the possibility of zero-shot translation~\cite{firat2016zero,johnson2017google,lakew2018comparison}. For example, a multilingual NMT model trained on $X \leftrightarrow$ English and $Y \leftrightarrow$ English parallel data is possible to translate between $X$ and $Y$ even if it has never seen the parallel data between $X$ and $Y$. \citeay{firat2016zero} achieve zero-resource translation by first training a multilingual model and generating a pseudo-parallel corpus for the target language pair via back translation, then fine-tuning on the pseudo-parallel corpus. Based on a fully shared multilingual NMT model, zero-shot translation shows reasonable quality without any additional steps \cite{johnson2017google}. Some designs can further improve the zero-shot translation in a fully shared multilingual NMT model: (1) introducing an attentional neural interlingua component between the encoder and decoder~\cite{lu2018neural}, (2) introducing additional terms to the training objective~\cite{arivazhagan2019missing,al2019consistency,pham2019improving}, and (3) correcting the off-target zero-shot translation issue via a random online back translation algorithm~\cite{zhang2020improving}. There are two important observations: (1) incorporating more languages may provide benefits on zero-shot translation~\cite{aharoni2019massively}, and (2) the quality of zero-shot between similar languages is quite good~\cite{arivazhagan2019massively}.


\subsection{Transfer learning}
A typical method of transfer learning for low-resource NMT is to first train an NMT model on some auxiliary (usually rich-resource) language pairs, which is called parent model, and then fine-tune all or some of the model parameters on a low-resource language pair, which is called child model~\cite{zoph2016transfer}. There are three main design aspects in transfer learning: (1) how to select the auxiliary language, (2) how to design the joint vocabulary between the auxiliary and low-resource languages, and (3) how to fine-tune the model for low-resource languages.

\paragraph{Auxiliary language selection.} A common approach is to select rich-resource languages as auxiliary languages~\cite{zoph2016transfer}, where the languages share the same/similar language family or typology with the given low-resource language performs better~\cite{nguyen2017transfer,dabre2017empirical}. LANGRANK is a framework to automatically detect the optimal auxiliary language based on typological and corpus statistical information~\cite{lin2019choosing}.

\paragraph{Joint vocabulary design.} A shared vocabulary including learned sub-words of the auxiliary and the desired low-resource language pairs is commonly used~\cite{nguyen2017transfer,kocmi2018trivial,gheini2019universal}. The shared vocabulary is often built by Byte Pair Encoding (BPE)~\cite{sennrich2015neural} and Sentencepiece~\cite{kudo2018sentencepiece}. However, the shared vocabulary is not suitable for transferring a pre-trained parent model to languages with unseen scripts in the vocabulary. To address this problem, \citeay{kim2019effective} propose to learn a cross-lingual linear mapping between the embeddings of the unseen language and the bilingual parent model.

\paragraph{Fine-tuning.} One simple method of fine-tuning is to use a parent model on one rich-resource language to initialize the child model and then fine-tune all the parameters on the low-resource language pair~\cite{zoph2016transfer}. Compared to fine-tune only on the desired low-resource language pair, a multistage fine-tuning procedure performs better, where the pre-trained parent model is first fine-tuned on a mixed corpora of the rich-resource and low-resource languages, and then fine-tuned on the desired low-resource language pair~\cite{dabre2019exploiting}. Some parameters can be fixed while fine-tuning, where \citeay{bapna2019simple} fix the parameters of the parent model and add light-weight residual adapters when fine-tuning. Moreover, besides using a bilingual parent model, a multilingual parent model can also be used, which enjoys two main advantages. First, a low-resource language can benefit from multiple auxiliary languages. Second, considering the limited model capacity of a multilingual model, fine-tuning may force the model to focus on the desired low-resource language, and thus improve the performance. \citeay{neubig2018rapid} compare different settings when fine-tuning a low-resource NMT model from a multilingual model on many-to-English direction, and come up with the conclusions: (1) Warm start, where the parent model is trained with both auxiliary languages and low-resource language, is better than cold start, where the parent model is trained only on auxiliary languages; (2) Fine-tuning from a universal model containing dozens of languages outperforms fine-tuning from a model with one similar auxiliary language; (3) Fine-tuning with the data of the low-resource language and one similar rich-resource language outperforms fine-tuning with only low-resource language data. In addition, \citeay{tan2019study} suggest warm start for many-to-one setting and cold start for one-to-many setting.


\subsection{Pivot translation}
In pivot-based approaches, a pivot language, which is usually a rich-resource language, is selected as a bridge. Then, the source-pivot and pivot-target corpora and model can be exploited to build the source-target translation.

There are mainly three ways to take advantage of pivot language. The first approach is to train the source-pivot and pivot-target models and directly combine them as a source-pivot-target model~\cite{cheng2017joint,ren2018triangular}. Second, another widely used method is to train the source-target model by pseudo-parallel data, which is generated with the help of the pivot language. \citeay{zheng2017maximum} translate the pivot language in a pivot-target parallel corpus to source language by a pivot-source NMT model, while \citeay{chen2017teacher} build the pseudo-parallel corpora by the source-pivot corpus and pivot-target model. Besides the parallel corpus, the monolingual data of source and target languages can also be used to generate pseudo-parallel corpora~\cite{karakanta2018neural,he2019language}. Moreover, the abundant monolingual data on pivot language can also be utilized to get the source-target pseudo-parallel corpora~\cite{currey2019zero}. Third, leveraging the parameters of source-pivot and pivot-target models is also one way to utilize the pivot language. \citeay{kim2019pivot} transfer the encoder of source-pivot model and the decoder of the pivot-target model to the source-target model. \citeay{ji2020cross} pre-train a universal encoder for source and pivot languages based on cross-lingual pre-training~\cite{conneau2019cross}, and then train on pivot-target parallel data with part of the encoder frozen. Pivot languages selection is critical in pivot-translation, which greatly influences the translation quality. In most cases, one pivot language is selected based on prior knowledge. There also exists a learning to route (LTR) method to automatically select one or several pivot languages to translate via multiple hops~\cite{leng2019unsupervised}.

\subsection{Summary and Discussions}
Both multilingual training and transfer learning are good ways to learn from auxiliary languages. In multilingual training, a low-resource language is trained with auxiliary languages from scratch, while in transfer learning, an existing translation model is fine-tuned on a low-resource language. Multilingual training and transfer learning can be combined by fine-tuning from a multilingual model. Pivot translation can be used when the translation path from a source language to a target language can be linked with one or several pivot languages, where each language pair on the path has sufficient training data to ensure high-quality translation. In practice, the methods in Section~\ref{sec:mono} and ~\ref{sec:multi} can be combined to further improve the translation accuracy on low-resource languages. For example, one can first train a multilingual NMT model, and then fine-tune it to a low-resource language with iterative back and forward translation.

\section{Exploiting Multi-Modal Data}
\label{sec:multimodal}
The parallel data in other modality is also useful for NMT, such as image, video, speech, etc. \citeay{chen2019words} and \citeay{huang2020unsupervised} built a pseudo parallel corpus by generating captions of the same image in both source and target languages via pre-trained image captioning models. In addition, the image caption/description and translation tasks can be jointly learned to incorporate image information~\cite{luong2015multi,specia2016shared,chen2018zero}. Moreover, the image data can be utilized by introducing an additional image component (e.g., encoder, decoder or attention) into the NMT model and aligning the two languages with the corresponding image in the latent space~\cite{huang2016attention,libovicky2016cuni,su2019unsupervised,nakayama2017zero,calixto2017doubly,elliott2017imagination,zhou2018visual,lee2017emergent}. Currently, the application of using image-text parallel data on NMT is limited, since such image-text data is always hard to collect for low-resource languages. One potential data source to build new image-text dataset is the images and corresponding captions on websites (e.g., Wikipedia and news pages). For the languages with only speech but no text scripts, speech data can be leveraged to develop the translation capability~\cite{zhang2020uwspeech}.


\section{Datasets}
\label{sec:data}
\begin{table}[h]
\begin{tabular}{llll}
\toprule
Dateset & Type & \#Language & Size \\ \midrule 
Wikipedia & mo & 300+ & $\sim55$M documents \\
CommonCrawl & mo & 150+ & Billions of URLs \\
CC-100 & mo & 100+ & $\sim0.5$B sents/lang \\
JW300 & bi & 300+ & $\sim0.1$M sents/pair \\
CCAligned & bi & 137 & $\sim0.3$M sents/pair \\
CCMatrix & bi & 80 & $\sim1+$M sents/pair \\
WikiMatrix & bi & 85 & $\sim0.1$M sents/pair \\\bottomrule
\end{tabular}
\caption{List of datasets, where mo and bi stand for monolingual and bilingual data, sents/lang and sents/pair stand for the number of sentences in a language and language pair, respectively.}
\label{tab:dataset}
\end{table}
Data is critical for low-resource NMT. In this section, we introduce some corpora that are widely used in low-resource NMT, as shwon in Tab. \ref{tab:dataset}. Wikipedia\footnote{https://www.wikipedia.org/} and Common Crawl\footnote{http://commoncrawl.org/} contain abundant monolingual data, where Wikipedia covers more than 300 languages and Common Crawl contains billions of web pages crawled from the Internet. CC-100~\cite{conneau2020unsupervised,wenzek2020ccnet} is a monolingual corpus covering 100+ languages processed from Common Crawl. JW300~\cite{agic2019jw300}, CCAligned~\cite{el2020massive}, CCMatrix~\cite{schwenk2019ccmatrix} and WikiMatrix~\cite{schwenk2020wikimatrix} extract parallel sentences from monolingual data, where JW300 is from the website jw.org, CCAligned and CCMatrix are aligned from Common Crawl, and WikiMatrix is based on Wikipedia. Moreover, OPUS~\cite{tiedemann2012parallel} and HuggingFace\footnote{https://huggingface.co/} provide a collection of open source corpora, which makes it much convenient to collect data from multiple data sources.
\section{Conclusion and Future Directions}
\label{sec:con}
In this paper, we provided a literature review for low-resource NMT. Different techniques are classified based on the type of auxiliary data: monolingual data from the source/target languages, data from other languages, and multi-modal data. We hope this survey can help readers to better understand the field and choose appropriate techniques for their applications. 

Though lots of efforts have been made on low-resource NMT as surveyed, there still remain some open problems: 
\begin{itemize}
    \item In multilingual and transfer learning, how many and which auxiliary languages to use is unclear. LANGRANK~\cite{lin2019choosing} trains a model to select one auxiliary language. Intuitively, using multiple auxiliary languages may outperform only one, which is worth exploration.  
    \item Training a multilingual model including multiple rich-resource languages is costly. Transferring a multilingual model to an unseen low-resource language is an efficient approach, where the challenge is how to handle the new vocabulary of the unseen language.
    \item How to efficiently select pivot language(s) is important but has not been well investigated.
    \item Bilingual dictionary is useful and easy-to-get. Current works focus on taking advantage of bilingual dictionary on the source and target language. It is also possible to use bilingual dictionary between a low-resource language and auxiliary languages in multilingual and transfer training.
    \item In terms of multi-modality, speech data has potential to boost NMT, but such studies are limited. For example, some languages are close in speech but different in script (e.g., Tajik and Persian).
    \item Current approaches have made significant improvements for low-resource languages that either have sufficient monolingual data or are related to some rich-resource languages. Unfortunately, some low-resource languages (e.g., Adyghe and Xibe) have very limited monolingual data and are distant from rich-resource languages. How to handle such languages is challenging and worth further studies.
\end{itemize}



\bibliographystyle{named}
\bibliography{ref}

\end{document}